\documentclass[10pt,twocolumn,letterpaper]{article}

\usepackage{cvpr}              

\usepackage[dvipsnames]{xcolor}

\definecolor{cvprblue}{rgb}{0.21,0.49,0.74}
\usepackage[pagebackref,breaklinks,colorlinks,citecolor=cvprblue]{hyperref}

\usepackage{times, graphicx, makecell, float}
\usepackage{latexsym}
\usepackage{hyperref}
\usepackage[T1]{fontenc}
\usepackage[utf8]{inputenc}
\usepackage{booktabs}
\usepackage{multirow}
\usepackage{tabularx}

\def\paperID{*****} 
\def\confName{CVPR}
\def\confYear{2024}

\title{ColorFoil: Investigating Color Blindness in Large Vision and Language Models}

\author{Ahnaf Mozib Samin\\
University of Malta\\
Msida, Malta\\
{\tt\small ahnaf.samin.22@um.edu.mt}
\and
M. Firoz Ahmed\\
Shahjalal University of Science and Technology\\
Sylhet, Bangladesh\\
{\tt\small mfiroz.sust@gmail.com}
\and
Md. Mushtaq Shahriyar Rafee\\
Metropolitan University\\
Sylhet, Bangladesh\\
{\tt\small rafee.sust@gmail.com}
}

\begin{document}
\maketitle
\begin{abstract}

With the utilization of Transformer architecture, large Vision and Language (V\&L) models have shown promising performance in even zero-shot settings. Several studies, however, indicate a lack of robustness of the models when dealing with complex linguistics and visual attributes. In this work, we introduce a novel V\&L benchmark - ColorFoil, by creating color-related foils to assess the models' perception ability to detect colors like red, white, green, etc. We evaluate seven state-of-the-art V\&L models including CLIP, ViLT, GroupViT, and BridgeTower, etc. in a zero-shot setting and present intriguing findings from the V\&L models. The experimental evaluation indicates that ViLT and BridgeTower demonstrate much better color perception capabilities compared to CLIP and its variants and GroupViT. Moreover, CLIP-based models and GroupViT struggle to distinguish colors that are visually distinct to humans with normal color perception ability.
\end{abstract}    
\section{Introduction} \label{intro}

Vision and language models (V\&L) have exhibited improved performance for many V\&L tasks in recent years \cite{lu2019vilbert, su2019vl, chen2020uniter, li2020unicoder, radford2021learning,dou2022empirical}. Thus, the current paradigm has now been shifting towards zero-shot learning, where models are evaluated without fine-tuning for specific tasks \cite{radford2021learning}. Large-scale V\&L models, in particular, show promise for task-independent zero-shot evaluation \cite{radford2021learning}.

Several studies have been conducted to perform comprehensive evaluations of V\&L models on a variety of tasks to identify their strengths and weaknesses \cite{agrawal2016analyzing, jabri2016revisiting, goyal2017making, shekhar2017foil, agarwal2020towards}. For instance, the VALSE evaluation benchmark has been proposed to assess the state-of-art V\&L models for challenging linguistic constructs \cite{parcalabescu2021valse}. Therefore, five distinct tasks, including existence, plurality, counting, relations, actions, and coreference, have been introduced. In this benchmark, foils are generated from the existing V\&L datasets for each of the tasks. A foil is referred to as a distractor or slightly incorrect example that is passed along with the correct example to the V\&L model to assess the model's ability to correctly distinguish them \cite{shekhar2017foil, parcalabescu2021valse}. Although the existing V\&L benchmarks like VALSE help the community to test the capabilities of V\&L models, there is still much work to be done to evaluate the robustness and generalizability of the models on numerous other tasks. It remains unknown how well the large V\&L models can perceive colors from the visual content.

Color perception requires a human-like understanding of visual content. Thus, by evaluating the V\&L models on color attributes, we can determine how closely the large V\&L models perceive colors to humans. A V\&L model can be biased towards detecting particular colors and perform poorly with others.  Therefore, it is essential to investigate it in order to improve the explainability and interpretability of the models. By assessing the V\&L models with their color-perception ability, we can ensure robustness in real-life applications.

In this study, we aim to shed light on the following research question: how well can the state-of-the-art large-scale V\&L models perceive color-related attributes, such as red, green, yellow, etc.? Our contributions are mainly twofold:

\begin{itemize}
  \item We introduce a novel V\&L benchmark \textbf{ColorFoil} by creating foils from the MS COCO and Flickr30k datasets \cite{lin2014microsoft, plummer2015flickr30k} to investigate how well the models perceive and identify the color-related attributes.
  \item We perform a comparison between seven of the state-of-the-art V\&L models including CLIP \cite{radford2021learning}, ViLT \cite{kim2021vilt}, ViT \cite{dosovitskiy2020image} and BridgeTower\cite{xu2022bridge} using our benchmark.
\end{itemize}

The outline of this paper is as follows. We provide a background study in Section \ref{background}. In Section \ref{construction}, we describe the process of constructing ColorFoil from the MS COCO dataset. Experiment setup is provided in Section \ref{exp_setup}. We report the results and discuss them in Section \ref{results}. In Section \ref{limitation}, we discuss the limitations of our work. Ethical considerations are provided in \ref{ethical}. A conclusion and future scope is presented in Section \ref{conclusion}.

\section{Background} \label{background}

\textbf{V\&L Models} The current state-of-art models are first pre-trained in a self-supervised way with a multi-task learning objective. The learning objectives can be predicting the masked texts or masked region in the images, determining whether or not the image and text corresponds, etc. The text and image input features can be concatenated together and passed to a Transformer encoder. This approach is known as single stream. Alternatively, the text and image inputs can be separately encoded to two different Transformers and then additional layers to merge them into multi-modal features.

\textbf{CLIP} Contrastive Language-Image Pre-training (CLIP) is a V\&L model that is pre-trained with 400M image-text pairs with a contrastive objective \cite{radford2021learning}. The model jointly trains a text encoder and an image encoder to maximize the cosine similarity of the image-text embeddings of real pairing while minimizing the cosine similarity of the embeddings of the incorrect pairings within a multi-modal embedding space. Each of the encoders are based on transformers. CLIP demonstrates the ability to perform zero-shot visual classification, object detection, and image generation tasks. 

\textbf{ViLT} Vision-and-Language Transformer (ViLT) is pre-trained using a Transformer with more than 4M images with two objectives such as image text matching and masked language modeling \cite{kim2021vilt}. The text embedding and the image features are concatenated into a sequence and then fed into the transformer. Thus, ViLT is a single stream model. ViLT achieves competitive or better performance than other V\&L models on downstream tasks while being 10 times faster due to simpler processing of visual inputs.

\textbf{BridgeTower} There is a visual encoder, a textual encoder and a cross-modal encoder with multiple lightweight bridge layers in the BridgeTower architecture \cite{xu2022bridge}. The top layers of the unimodal encoders and each layer of the cross-modal encoder are connected with the bridge layers, thus enabling extensive interactions at each layer of the cross-modal encoder. Each of visual, textual and cross-modal encoders is transformer-based encoders. The model is pre-trained with 4M images with two common objectives: masked language modeling and image text matching. The model is found to outperform in all downstream V\&L tasks with negligible additional computational cost.

\textbf{ViT} A Vision Transformer (ViT) is designed for image classification tasks, adapting the Transformer architecture from natural language processing \cite{dosovitskiy2020image}. It divides an image into fixed-size patches, linearly embeds each patch, and treats these embeddings as sequences akin to word tokens in text. Using self-attention mechanisms, the ViT captures global image context more effectively than convolutional networks, allowing for superior performance on large-scale image datasets. ViTs leverage transfer learning and pretraining for enhanced accuracy and efficiency.

\textbf{GroupViT} (Group Vision Transformer) is a variant of the Vision Transformer designed to improve efficiency and scalability in image classification tasks \cite{xu2022groupvit}. It enhances the standard ViT by introducing a group-wise processing mechanism, where the input image is divided into smaller groups of patches. Each group is processed independently through parallel self-attention layers, reducing computational complexity. The results from these groups are then aggregated to form a cohesive representation. GroupViT aims to retain the global context modeling capabilities of ViTs while optimizing resource usage, making it more suitable for large-scale and real-time applications.

\textbf{Related Work} Several V\&L tasks include visual question answering \cite{goyal2017making}, visual reasoning \cite{suhr2018corpus}, image retrieval \cite{plummer2015flickr30k}, etc. Foiling is an approach that slightly edits the original captions to evaluate the robustness of the V\&L models \cite{shekhar2017foil}. Similar to our work, \citet{shekhar2017foil} foiled the MS COCO dataset, and constructed the FOIL-COCO dataset. However, their work did not focus on the perception of colors of the V\&L models. Following the work of \citet{shekhar2017foil}, several studies have been performed that evaluated the V\&L models \cite{shekhar2019evaluating, gokhale2020mutant, bitton2021automatic, parcalabescu-etal-2021-seeing, rosenberg2021vqa}.
\section{Construction of the ColorFoil benchmark} \label{construction}

The ColorFoil benchmark is automatically derived from the MS COCO (Microsoft Common Objects in Context) and Flickr30k dataset, which serves as a resource for studying image understanding, object recognition, image captioning, and visual question-answering tasks \cite{lin2014microsoft, plummer2015flickr30k}. In the MS COCO dataset, textual annotations are provided solely for the train and validation (val) sets. To construct the ColorFoil, we obtain the images and annotations from the 2017 MS COCO validation set, resulting in a total of 5,000 image-text pairs. Among these instances, each of 2,511 pairs includes at least one word related to color. For Flickr30k dataset, we use the standard val and test sets to prepare the ColorFoil benchmark.

Our aim is to foil only the color name from the textual input, leaving the original image and the rest of the text input as it is. For example, given a caption like \textit{A \textbf{blue} bus driving down a street past a park.} We foil the color-related word, resulting in a modified sentence like - \textit{A \textbf{brown} bus driving down a street past a park.} If there are multiple color attributes in a caption, we foil all of them.

\begin{figure*}[!ht]
 \begin{center}
 \centering{\includegraphics[scale=0.27]{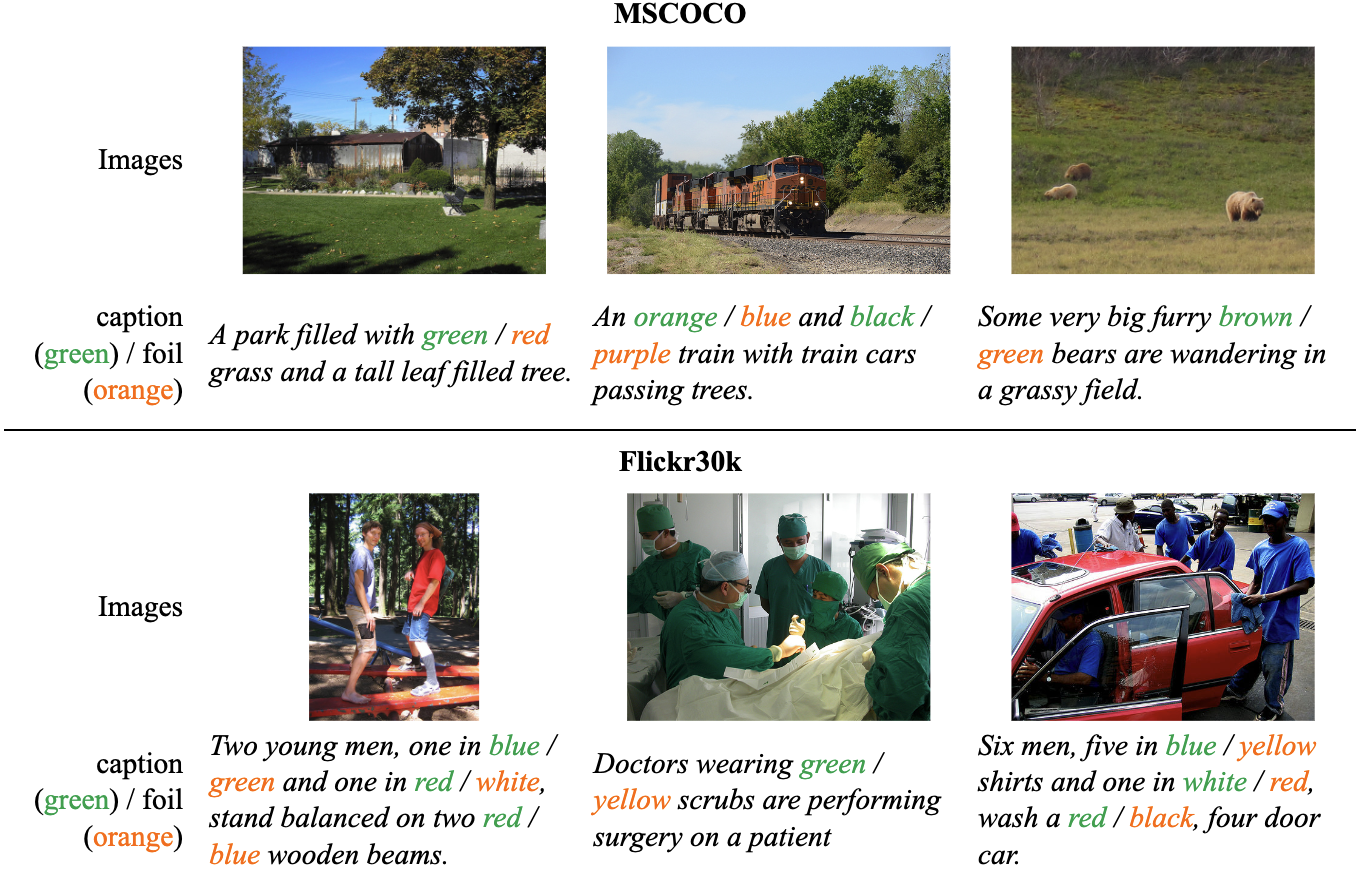}}
 \caption{Examples from the ColorFoil benchmark where color-related attributes in the original captions have been modified to different colors.}
\label{fig:examples}
\end{center}
\end{figure*}


We utilize the \textbf{webcolors 1.3} python package to determine whether a substring within a caption corresponds to a color \cite{pypi}. This package encompasses a total of 147 colors. Our filtering process involves excluding captions that lack color names and selecting solely those containing at least one color name. 

When replacing the original color name with a foiled alternative, we consider the most widely used colors. The chosen target colors for foiling consist of "blue", "black", "red", "pink", "yellow", "grey", "orange", "white", "green", and "brown." So, rather than utilizing the complete list of 147 colors from the \textbf{webcolors} package, we opt for a narrower selection of common colors for foiling. This decision is based on the fact that numerous colors in the package have limited practical usage (e.g. medium blue, mint cream, etc.). The target color for foiling is selected randomly from the 10 common colors. If the original color in the caption is one of the common colors, we randomly select any other common color for foiling except for the one found in the caption.

After excluding four instances of two-dimensional grayscale images due to compatibility issues with certain models, our resulting dataset comprises 2,507 pairs of RGB images along with their captions and foils from MSCOCO and 2500 pairs of RGB image-caption pairs from Flickr30k. To ensure data integrity, we conduct manual validation on a significant number of image-text pairs randomly selected from the benchmark and find no anomalies. Examples of original captions and corresponding foils are illustrated in Figure \ref{fig:examples}.
\section{Experimental Setup} \label{exp_setup}

We pass the original caption, foil as well as the corresponding image to a V\&L model. The model provides the logits for each of the caption and foil corresponding to the image. We take the softmax of the logits. Our hypothesis is that a model with a well-perceivable ability to distinguish colors is supposed to provide a higher probability for the original caption and a lower probability for the foil.

We evaluate all the models in a zero-shot setting. We utilize the HuggingFace transformer library to load the models \cite{wolf2019huggingface}. These models are chosen due to the fact that they represent different architectural variants. CLIP has a text encoder and an image encoder, which are jointly trained with a contrastive loss. ViLT is a single-stream model. BridgeTower contains multiple bridge layers that connect the uni-modal encoders with the cross-modal encoder.

The evaluation metric employed in our study is accuracy and F1-score, which are widely used in similar contexts. To elaborate, if the model accurately identifies the foil in comparison to the original caption, the accuracy of that particular example is incremented.

\begin{table*}[t]
\caption{\textbf{Experiment results.} We evaluate seven of the state-of-the-art V\&L models on the MS COCO and Flickr30k subsets from ColorFoil. Accuracy (\%) and F1-scores (\%) are reported. We conduct three experiments in which the models are presented different number of foils (modified caption) along with the original caption. The V\&L models tend to struggle in challenging conditions with more foils. BridgeTower and ViLT outperform other V\&L models including CLIP and its variants and GroupViT by a large margin.}
\centering
\setlength{\tabcolsep}{4pt}
\begin{tabular}{ccc|cc|cc|cc|cc|cc}
\hline\rule{0pt}{9pt}
\multirow{3}{*}{\textbf{Models}} & \multicolumn{4}{|c|}{\textbf{1 Foil}} & \multicolumn{4}{c|}{\textbf{2 Foils}} & \multicolumn{4}{c}{\textbf{4 Foils}}\\ \cline{2-13}\rule{0pt}{9pt} 
                       & \multicolumn{2}{|c|}{MSCOCO} & \multicolumn{2}{c|}{Flickr30k} & \multicolumn{2}{c|}{MSCOCO} & \multicolumn{2}{c|}{Flickr30k} & \multicolumn{2}{c|}{MSCOCO} & \multicolumn{2}{c}{Flickr30k}\\ \cline{2-13}\rule{0pt}{9pt}
                       & \multicolumn{1}{|c} {Accuracy}       & F1        & Accuracy        & F1          & Accuracy       & F1        & Accuracy        & F1          & Accuracy       & F1        & Accuracy        & F1          \\ \hline\rule{0pt}{9pt}
                       
ALIGN \cite{jia2021scaling}                  & \multicolumn{1}{|c}{86.03}          & 93.32      & 87.70           & 93.45        & 79.47          & 88.75      & 81.43           & 89.76        & 71.03          & 83.06      & 74.57           & 85.43        \\

AltCLIP \cite{chen2023altclip}                & \multicolumn{1}{|c}{84.89}          & 91.82      & 82.69           & 90.52        & 77.29          & 87.19      & 73.68           & 84.84        & 69.08          & 81.71      & 64.39           & 78.34        \\

BridgeTower \cite{xu2022bridge}           & \multicolumn{1}{|c}{97.31}          & 98.63      & 96.83           & 98.32        & 95.71          & 97.81      & 94.46           & 97.15        & 92.61          & 96.16      & 90.81           & 95.18        \\

CLIP \cite{radford2021learning}                 & \multicolumn{1}{|c}{84.42}          & 91.55      & 85.24           & 92.07        & 76.19          & 86.49      & 76.26           & 86.53        & 67.37          & 80.50      & 68.33           & 81.18        \\

CLIPSeg \cite{luddecke2022image}               & \multicolumn{1}{|c}{83.05}          & 91.09      & 82.01           & 90.12        & 74.00          & 85.42      & 72.97           & 84.37        & 64.56          & 78.45      & 63.07           & 77.35        \\

GroupViT \cite{xu2022groupvit}            & \multicolumn{1}{|c}{82.73}          & 91.67      & 81.64           & 89.89        & 73.10          & 83.98      & 71.77           & 83.57        & 63.80          & 77.89      & 62.12           & 76.63        \\

ViLT \cite{kim2021vilt}                 & \multicolumn{1}{|c}{95.69}          & 97.79      & 94.29           & 97.06        & 92.83          & 96.28      & 91.85           & 95.35        & 88.74          & 94.04      & 87.38           & 93.27    \\ \hline\rule{0pt}{9pt}  
\end{tabular}
\label{table: Model Comparison}
\end{table*}

\begin{table*}[h]
    \caption{Examples of caption-foil pairs for which the CLIP model wrongly choose the foils instead of the captions.}
    \label{table: clip_error}
    \centering
    \begin{tabular}{ c  l}
        \hline\rule{0pt}{10pt}
        \textbf{Type}  & \textbf{Example}\\
        \hline\rule{0pt}{9pt}
        caption &  A surfer is riding a wave in light \textbf{blue} water.\\
        foil &  A surfer is riding a wave in light \textbf{brown} water.\\
        \hline
        caption &  A man in \textbf{black} jersey pitching in baseball game.\\
        foil & A man in \textbf{red} jersey pitching in baseball game.\\
        \hline
        caption & A \textbf{red} traffic light at night next to a Christmas Tree.\\
        foil & A \textbf{white} traffic light at night next to a Christmas Tree.\\
        \hline
    \end{tabular}
\end{table*}

\section{Results and Discussion} \label{results}

Table 2 shows the performance of different V\&L models evaluated on the ColorFoil. All the models achieve much higher accuracy compared to a baseline random classifier with a 50\% accuracy. CLIP obtains 83.1\% accuracy while ViLT and BridgeTower get substantially higher accuracy of 95.6\% and 97.2\%, respectively on the 1-Foil experiment. It is worthwhile to mention that CLIP is pre-trained with 400M images, although this model is outperformed by both ViLT and BridgeTower pre-trained with only 4M images. BridgeTower architecture, which contains multiple bridges to make connections between the uni-modal encoders and the cross-modal encoder, achieves the highest accuracy.

The relatively poor performance of CLIP is also evident in its variants, including AltCLIP and CLIPSeg. While the ALIGN model outperforms CLIP, it still lags behind BridgeTower and ViLT. GroupViT, similar to CLIP, struggles to achieve high performance. This performance trend is consistent across both MSCOCO and Flickr30k datasets, reinforcing our observations. When presented with more foils alongside the original caption, the models exhibit performance degradation. Nonetheless, BridgeTower and ViLT maintain strong performance even under these challenging conditions with more foils.

We present several pairings of captions and foils for which the CLIP model incorrectly assigns higher probabilities to the foils (See Table \ref{table: clip_error}). These examples demonstrate that the CLIP model is unable to distinguish between blue-brown, black-red, and red-white pairs, despite the fact that they are visually distinct to most humans.
\section{Limitations} \label{limitation}

We consider the 10 most common colors for our foils. However, our choice of common colors is subjective and there might be other frequently used colors that are not present in our foils. 
\section{Ethical Considerations} \label{ethical}

Training V\&L models using images and corresponding texts that may contain gender bias, private data, or harmful content presents challenges in manual detection. To address this, we utilize the widely recognized MS COCO and Flickr30k datasets to create the ColorFoil benchmark, as it provides a reliable foundation \cite{lin2014microsoft, plummer2015flickr30k}.

Ensuring reproducibility is a crucial aspect of scientific research. To foster open research practices, we will make our code publicly accessible, allowing others to reproduce and verify our findings.
\section{Conclusion and Future Work} \label{conclusion}

In this work, we introduce a novel benchmark, ColorFoil, derived from the MS COCO and Flickr30k datasets, to assess the perception ability of the cutting-edge V\&L models to detect colors. To this end, we foil the colors from the original captions and feed both caption and foil along with the corresponding image to the model to observe whether it can provide a higher probability for the caption or not. Seven state-of-the-art V\&L models, including CLIP, ViLT, ViT, and BridgeTower, have been benchmarked using the ColorFoil. While all models outperform a random classifier, ViLT and BridgeTower are much more capable to perceive colors compared to CLIP and ViT. This intriguing finding is seen using both MS COCO and Flickr30k datasets, which strengthens our analysis. 

As part of our future work, we would like to evaluate the robustness of V\&L models on additional tasks by constructing foils that swap gender (man -> woman), size (small -> large), emotions (smiling -> crying), and sentence negation (playing football -> not playing football), etc.
{
    \small
    \bibliographystyle{ieeenat_fullname}
    \bibliography{main}
}


\end{document}